\documentclass[10pt, conference, compsocconf]{IEEEtran}

\usepackage{cite}
\usepackage{epsfig}
\usepackage{flushend}
\usepackage{mathtools}
\usepackage{subfigure}
\usepackage{calc}
\usepackage{amssymb}
\usepackage{amstext}
\usepackage{amsmath}
\usepackage{amsfonts}
\usepackage{amsfonts}
\usepackage{amsthm}
\usepackage{multicol}
\usepackage{pslatex}
\usepackage{apalike}
\usepackage{dsfont}
\usepackage{graphicx}
\usepackage{bm}

\graphicspath{{images/}}

\begin{document}
\title{Anomaly Detection with\\HMM Gauge Likelihood Analysis}

\author{\IEEEauthorblockN{Boris Lorbeer, Tanja Deutsch, Peter Ruppel, Axel K\"upper}
\IEEEauthorblockA{Service-centric Networking\\
Technische Universit\"at Berlin\\
Berlin Deutschland\\
lorbeer@tu-berlin.de, tanja.deutsch@tu-berlin.de, peter.ruppel@tu-berlin.de, axel.kuepper@tu-berlin.de}
}

\maketitle

\begin{abstract}
This paper describes a new method, {\em HMM gauge likelihood analysis}, or GLA,
of detecting anomalies in discrete time series using Hidden Markov Models and
clustering. At the center of the method lies the comparison of subsequences. To
achieve this, they first get assigned to their Hidden Markov Models using the
Baum-Welch algorithm. Next, those models are described by an approximating
representation of the probability distributions they define. Finally, this
representation is then analyzed with the help of some clustering technique or
other outlier detection tool and anomalies are detected. Clearly, HMMs could be
substituted by some other appropriate model, e.g.\ some other dynamic Bayesian
network. Our learning algorithm is unsupervised, so it doesn't require the
labeling of large amounts of data. The usability of this method is demonstrated
by applying it to synthetic and real-world syslog data.
\end{abstract}


\section{Introduction}
\label{sec:introduction}

\noindent The detection of anomalies in data is currently one of the most
important practical applications of unsupervised learning. It helps companies
to understand their data better, to find hidden flaws in complex systems, and
to react early to the emergence of unexpected situations. The assumption here
is that anything that is anomalous is potentially indicating some kind of
failure, malicious behavior, or otherwise exceptional incident that needs
attending to. While anomalous patterns can occur in all kinds of data, in this
paper we are exclusively dealing with anomaly detection in time series. One
important example of time series data in a complex system would be the logging
data in a computer network. Today the computer networks of large companies can
easily consist of several thousands of machines, each one running a wide range
of applications that often use the logging framework syslogd to log their
status and important events. This is, of course, a valuable source for
diagnostics and predictive maintenance.

The thorough analysis of large amounts of data is usually hampered by mainly
two problems: first, the amount of data is often too huge for humans to read
through, and second, the experts which can interpret the data are usually very
hard to find. Thus, it would be very helpful, if this analysis could be
automated, at least partially. One approach often used is that of anomaly
detection: use machine learning to learn models that describe the normal
behavior and then label those patterns, that cannot be described well by the
learned models, as anomalous. The advantages of anomaly detection are that
large amounts of data can be processed automatically and that no experts are
necessary. Moreover, anomaly detection algorithms can often detect strange
patterns in data that look innoxious even to an expert. Also, the algorithms
used in this paper are unsupervised, i.e.\ they don't need labeled data for
learning, which for large data is in general very difficult to obtain.  The
disadvantage is, that it is often difficult to determine the threshold of how
anomalous an event has to be for being noted as an anomaly, and that depending
on this threshold, we might obtain too many false positives or too many false
negatives. And, of course, there is always the possibility of rare but harmless
events, as well as critical events that have a high frequency; anomaly
detection cannot help in those situations and other methods have to be applied.

There are many different types of anomalies possible, as there are many
different types of patterns that data can contain, and violations of those
patterns can be considered anomalies. One very simple pattern in time series
data that could be exploited is the simple presence of certain events. And if
some events are usually not present, i.e.\ have a low frequency, their
occurrence would then be considered an anomaly. Thus, one conceptually very
easy anomaly detection method is the frequency analysis of events. We specify a
threshold for the frequency, and any event frequency below this threshold is
considered an anomaly and this rare event should be examined more closely. The
threshold itself depends on the given situation and should be adapted
accordingly. So in this case, the ``machine learning'' part simply consists in
recording the frequencies of events and then applying a threshold.

Other patterns, that could be used for time series anomaly detection are the ratios,
or more general linear dependencies of event counts, for example within a
certain time window. That means, contrary to the example above, here we don't
just consider individual events but rather aggregations of events. One way of
implementing this is by using PCA on event counts \cite{xu2010detecting}.

Yet another type of pattern is the consecution of events. Thus, while the
pattern in this case also consists of aggregations of events as the ratio
pattern above, now also the order of events is relevant. So, for example, if
our event time series consists of two events in the form (A, B, A, B, A, B,
...) and then suddenly there would appear the sequence (A, A, A, B, B, B), the
ratio anomaly detection would not be triggered, but the order anomaly detection
would. This variant has been investigated in many papers, and it is also the
focus of the current paper.

The method that is described in this paper could be delineated shortly as
follows: We consider one long time series of events and cut it into a series of
smaller sequences that are all of the same length, e.g.\ 20 events long. To
describe them properly, we map them to their Hidden Markov Models (HMM). Two
sequences are then considered similar if their HMMs are similar.  Since HMMs
are non-identifiable, comparing them is nontrivial. We consider them as
probability distributions over the space of event sequences. So we take a few,
e.g.\ 10, fixed event sequences and compute their probabilities wrt.\ the HMMs,
creating a vector of probabilities for each HMM. If those vectors are similar
for two HMMs, it is likely that the HMMs have similar probability distributions
and thus are similar themselves. This, in turn, means, that their original event
sequences are similar. Thus, we have a nontrivial feature engineering scheme,
that maps sequences to vectors of probabilities that can then be searched for
outliers, e.g.\ by clustering algorithms that detect outliers like for instance
HDBSCAN or simply by visually inspecting the two-dimensional t-SNE projection.

We demonstrate this method with synthetic data as well as real-world syslog
data.

\section{\uppercase{Related Work}}
Here we give a very short and rather incomplete review of the literature on
anomaly detection for time series data with a finite value space. For a more
comprehensive overview, see e.g. \cite{chandola2012anomaly}.

The approach most often used in anomaly detection with machine learning is to
learn a model from the normal patterns and then to declare all patterns, to
which the model assigns very low probability, to be anomalous.

Some authors partition the time into intervals and consider histograms of the
frequencies of the events in each window, which results in a collection of
histograms. Since the value (event) space is finite, those histograms are
points in a finite-dimensional vector space. This collection of points can now
be approximated with PCA and points very far away from the PCA subspace will be
labeled anomalous. This idea is exploited for example in
\cite{xu2010detecting}, \cite{xu2009online}, and \cite{xu2010experience}. In
those papers, neither the sequential ordering inside the windows nor the order
of the windows themselves is used, only the linear relations of the frequencies
of nearby events are taken into account.

If we want to include the order of events in our analysis, we enter the realm
of time series analysis. One very powerful tool in time series analysis are
Hidden Markov Models (HMMs) and so it is not surprising that they have been
used a lot for this kind of anomaly detection. There are many different
variants in which HMMs can enter the scene. The most popular approach is to
compute an HMM for the normal data and then compute for all new data sequences
their probabilities wrt.\ this HMM. If the probability of the sequence is too
small, i.e.\ is below a carefully chosen threshold, the sequence is labeled as
anomalous. This method can be found for example in \cite{chandola2012anomaly},
\cite{lane2003empirical}, \cite{joshi2005investigating},
\cite{khreich2009combining}, \cite{khreich2010iterative},
\cite{lane2003empirical} or \cite{wang2004modeling}. A similar idea can be
found in \cite{dorj2013bayesian}, which replaces HMMs with Bayesian HMMs. The
problem with all those techniques is that they are not very good in modeling
multiple types of sequences. They work best if there is just one main
sequential structure present in the normal data. However, we are interested in
situations of long sequences that can display several different patterns and it
turned out that this method does not give usable results.

HMMs are often interpreted as finding, to the given observed sequence, an
underlying sequence of ``hidden'' states, which are thought of as causing the
observations. Often, the HMM is chosen to have fewer values in the hidden
states than in the observations, which makes the hidden sequence less complex
and easier to compute with. One interpretation could be that the observations
are the states with added noise. This point of view leads to the idea of
replacing the observed sequence with the sequence of hidden states and then
doing anomaly detection on this one (for example by comparing them to a
database of labeled sequences, using an appropriate similarity measure). For
work along this line of reasoning, see e.g.\ \cite{chandola2012anomaly} or
\cite{chandola2008comparative}, and references therein. However, this method,
too, suffers from the drawbacks described before and is not usable for our
purposes.

One problem with HMMs is that with longer sequences with many states,
computations become slow and sometimes lead to arithmetic underflow. In
\cite{florez2005efficient} they developed an iterative version of HMMs that
alleviates those problems.

Another problem is the choice of the number of hidden states. Most often, this
parameter is simply tuned according to what works best in the given situation.
A more principled approach is taken by \cite{khreich2009combining}, which
choose a combination of HMMs with different state counts by using the ``Maximum
Realizable ROC'' technique. Then again, the anomalies are detected via a
carefully chosen probability threshold. An extension of this scheme can be
found in \cite{khreich2010iterative}. One further approach to anomaly detection
that uses combinations of HMMs is demonstrated in \cite{yamanishi2005dynamic}
which uses mixtures of HMMs. The machinery described is quite complex. However,
the clustering described there did not work well with our data. We think that
is due to two problems: first, the number of components needs to be provided to
the algorithm and the dynamic selection of this number described in the paper
was not very robust. Second, the range of possible shapes of clusters is
bounded and cannot be as arbitrary as with e.g.\ HDBSCAN.

\clubpenalty10000
Of course, today, with the rise of deep learning \cite{goodfellow2016deep} in
almost all fields of machine learning, there are also applications of this new
paradigm to anomaly detection. Some work has been published in this area in the
last few years, but we will not go into detail here and only mention two
general approaches. One considers anomaly detection simply as a classification
problem (i.e. is only applicable to labeled data), and can thus apply one of
the standard deep neural network architectures for supervised learning. E.g.\
in \cite{javaid2016deep}, they train a deep sparse autoencoder network. Another
approach does anomaly detection via time series prediction. It learns Recurrent
Neural Networks (RNNs) \cite{sutskever2013training} for the event sequences,
computes the probability distribution over all possible extrapolations for the
next few steps and labels sequences that have low probability with respect to
this distribution as anomalous. Quite a number of papers follow this scheme,
see e.g.\ \cite{malhotra2015long}, \cite{chauhan2015anomaly}, or
\cite{shipmon2017time}, to name only three. A clear advantage of Recurrent
Neural Networks, especially Long Short Term Memory (LSTM) models, see
\cite{hochreiter1997long}, is that they can much better remember past events
than HMMs can. One clear disadvantage of deep neural networks is the often
large amount of effort necessary to get the network to converge to a good model.
Since anomaly detection via prediction can be done with any prediction method,
classical or deep learning, it might be interesting for us to compare the
prediction performance of all kinds of prediction models. This has been done
several times, for recent results see e.g.\ \cite{makridakis2018statistical},
the NN3 competition \cite{crone2005automatic}, and the M4 competition
\cite{makridakis2018m4}.
In general, deep learning seems like a promising direction for anomaly
detection in time series, especially if long term correlations between events
are present.

\section{HMM gauge likelihood analysis}
We are concerned with the problem of finding anomalies in discrete time series.
In this paper, we consider only batch processing: first, the time series data
is collected, let's say for one whole day, and then this complete batch of data
is processed together. The scenario is that of a long time series, i.e.\ a
continuous stream of discrete data, that most of the time behaves normal but at
rare occasions digresses from that normal pattern of behavior. Our task is to
find those anomalies since they are likely to be of importance e.g.\ for
predictive maintenance. One example would be the syslog events written by a
given application. As long as the application always logs the same sequences of
events, everything is fine. However, if this pattern suddenly changes, we would
like to know about this change.

The first point to make is that we will not consider the complete time series
$T$ as a whole, but rather subsequences $t_k$ thereof. To be more precise: for
$i,j\in\mathds N, i \le j$, let the index set $\bm I_{i:j}$ be the set of
integers $\{i,i+1,\ldots,j\}$. Moreover, let $\mathfrak E$ be the set of
possible events we can observe. Then we consider all the subsequences $t^n_k$
of length $n$ of the total sequence $T$ of length $N$ that are obtained by {\em
sliding a window of size $n$ with a step size $m$} over the domain of $T$,
i.e.:
\begin{equation}\begin{split}
&T: \bm I_{1:N} \to \mathfrak{E}\\
&t^n_k: \bm I_{1:n} \to \mathfrak{E},\; n\in\mathds N,\;n\le N,\quad k = 1, 2, \ldots, \lfloor(N+m-n)/m\rfloor\\
&t^n_k(l) := T(l+(k-1)m),\quad l\in \bm I_{1:n},
\end{split}\end{equation}

\noindent where we often omit the superscript $n$. Note that at the end of $T$
we don't keep windows that are incomplete (in case $N-n$ is not dividable by
$m$). The window size $n$ should be a small multiple of the expected maximal
autocorrelation distance, so the model has a chance to learn those
correlations. As the window shift size $m$ we usually choose half the window
size: $n/2$.

It is those subsequences $t_k$ that we want to compare. Usually, most are
similar to each other and some are different and the latter are the anomalies.
The common length $n$ of those subsequences should not be too small, such that
they will still contain a sufficient amount of most of the behavioral patterns.
On the other hand, because of computational constraints like arithmetic
underflow and computation time, the length cannot be too large. Thus, this
parameter has to be tuned according to the situation at hand.

Our next step is to find a way to compare the subsequences $t_k, k=1, \ldots,
K$. Unfortunately, comparing sequences to each other is not trivial. We could,
for example, compare two sequences $t_r = (t_{ri})_{i=1}^n$ and $t_s =
(t_{si})_{i=1}^n$ by taking the Euclidean distance between those two vectors:
$d_E(t_r, t_s) = \sqrt{\sum_{i=1}^n|t_{ri}-t_{si}|^2}$. But this measure has
several flaws. For example, this would mean that two sequences that just differ
by a simple time shift could be considered completely different. There are
several possible approaches to measuring the similarity of sequences that avoid
this and similar problems. Our choice in this paper is to compare two sequences
by comparing the HMMs that have been learned from them using the Baum-Welch
algorithm. HMMs are an appropriate model in this situation since they are
simple enough to be tractable, yet sufficiently versatile to capture the
idiosyncrasies of the sequences they have been fitted to. This has been shown
by numerous applications of this popular model. E.g., they are, contrary to
ordinary Markov chains, capable of detecting longer lasting autocorrelations.
For a good introduction to HMMs, see e.g.\ \cite{rabiner1989tutorial},
\cite{bishop2006PRML}, or \cite{prince2012computer}.

To assign a HMM model to each sequence $t_k$, we have to choose a number of
hidden states. Again, this needs to be tuned for the given problem. The trade
off here is similar to that for the length of the sequence: large numbers of
hidden states on the one hand provide more powerful models, which for example
can remember further back in the past, but on the other hand lead to high
computational costs.

Unfortunately, HMMs cannot be compared just by comparing the model parameters.
Indeed, it is well known that HMMs are {\em non-identifiable}, i.e.\ there are
different sets of model parameters that define equivalent models. To understand
this, first note that statistical models such as HMMs assign a probability to
each possible data input, in our case to each possible sequence of the same
length $n$ as the HMM. More formally: Let $n$ be the chosen length of the event
sequences. Also, recall that $\bm I_n$ is the index set containing the first
$n$ natural numbers $\bm I_n := \{1, \ldots, n\}$, and that $\mathfrak{E}$ is
the set of all possible events that can happen. Then:
\begin{equation}
\mathcal{S}_n = \{s| s:\bm I_n \to \mathfrak{E}\}
\end{equation}

\noindent is the set of all possible event sequences of length $n$ with event
space $\mathfrak{E}$. Of course, our original subsequences $t_k$ are also
contained in this set: $t_k\in\mathcal{S}_n, k=1, \ldots, K$. Further, let
$\theta$ be the vector containing all the parameters of an HMM, and let
$h(\theta)$ be the belonging HMM. Then each HMM $h(\theta)$ of length $n$
defines a probability distribution function (PDF) $p_{h(\theta)}$ over
$\mathcal{S}_n$:
\begin{equation}
p_{h(\theta)}:\mathcal{S}_n\to[0, 1],\quad\sum_{s\in\mathcal{S}_n} p_{h(\theta)}(s) = 1.
\end{equation}

\noindent Now, for any practical purposes, two HMMs $h(\theta_1)$ and
$h(\theta_2)$ are indistinguishable if they define the same PDF over
$\mathcal{S}_n$:
\begin{equation}
h(\theta_1) = h(\theta_2) \quad \coloneqq \quad \forall_{s\in\mathcal{S}}\;\; p_{h(\theta_1)}(s) = p_{h(\theta_2)}(s).
\end{equation}

\noindent Indeed, they should be considered equal, since there is no experiment
that can prove them different. In general, models with the property that
different model parameters $\theta_1$ and $\theta_2$ can lead to the same
probability distribution are called non-identifiable. Now, as mentioned above,
HMMs happen to have this property, i.e., with the above notation:
\begin{equation}
\exists_{\theta_1, \theta_2, \theta_1\ne\theta_2}\quad h(\theta_1) = h(\theta_2).
\end{equation}

\noindent One very simple example is that we can single out two states and then
switch in the transition matrix the belonging two rows and columns, and in the
emission matrix the two rows belonging to the two states. This will not change
the probability distribution but the parameters will in general change. This
can, of course, be done with any arbitrary permutation of states. Note,
however, that there are many more ways in which identifiability is violated in
HMMs, not just via permutations.  For more background on non-identifiability of
HMMs and its algebraic geometrical structure see e.g.\
\cite{watanabe2001algebraic} and \cite{hosino2005stochastic}.  The important
implication for our purposes is that the parameters of HMMs cannot be used to
compare them since models that are similar or equivalent can have very
different sets of parameters.

The above explanation takes the approach of viewing statistical models, and in
our case HMMs in particular, as probability distributions over the data. And
this view is also what leads us to the feature map that we choose in our
approach. Ideally, we would take as features the probability distributions
$p_h(\theta)$ themselves. But, of course, comparing probability distributions
is not realistic because of the large size of $\mathcal{S}_n$ even for
moderately sized $n$ and $\mathfrak{E}$. Thus, instead of comparing the
probabilities of each and every single sequence $s\in\mathcal{S}_n$, only a
small number $m$ of them is selected. Those chosen sequences then define a map
$\ell$ from the set of HMM models $\mathcal{H}$ to $\mathds{R}^m$. More
formally: Let $\mathcal{G}$ be the set of sequences chosen to compare HMMs:
\begin{equation}
\mathcal{G} = \{g_1, \ldots, g_m\}, \quad g_i\in\mathcal{S}_n,\;i=1,\ldots,m.
\end{equation}

\noindent Then the map $\ell$ is defined as follows:
\begin{equation}
\begin{split}
&\ell: \mathcal{H} \to \mathds{R}^m,\\
&\ell(h) := (p_h(g_1), \ldots, p_h(g_m))\in\mathds{R}^m.
\end{split}
\end{equation}

The sequences $g_i\in\mathcal{G}$ are the basis for comparing the HMMs,
i.e.\ they gauge our comparisons. Because of this, we will refer to those
sequences as {\em gauge sequences}.

Thus, for the comparison of two models $h(\theta_1)$ and $h(\theta_2)$, we
don't use their parameters $\theta_1$ and $\theta_2$ but rather their images
under $\ell$ in $\mathds{R}^m$. As far as the choice of gauge sequences is
concerned, we investigated two approaches. One was choosing a couple of
sequences at random. The number of random gauge sequences was not very
important as long as it was not below five. The other approach was to use the
original sequences themselves. In our experiments, the latter approach showed
slightly better cluster separation properties, while requiring more computation
time.

To summarize, instead of comparing the sequences directly, we use a feature map
that assigns each sequence to a point in an $m$-dimensional cube $[0,1]^m$
where $m$ is the number of gauge sequences, and then analyze those image points
instead. To detect outliers in this feature space we first apply t-SNE
\cite{maaten2008visualizing} to map the points into $\mathds R^2$ and then use
HDBSCAN \cite{campello2013density}. It has turned out that using t-SNE before
clustering gives superior results. Moreover, the projection to a
two-dimensional space allows for easy visualization.

We refer to this method of outlier analysis as {\em HMM gauge likelihood
analysis}, or GLA for short.

\section{Application to synthetic and real data}
We evaluated the method on synthetic data as well as real-world data. The
results on real-world data were compared with those of a popular deep learning
anomaly detection method.

\paragraph{Synthetic data}
We have created 60 sequences of length 20, each consisting of the subsequence
(A, B, C, D), concatenated five times with small noise added by randomly
selecting two elements from each sequence that were altered randomly. Next, we
added two anomalous sequences, one which had the order of events reversed,
i.e.\ the sequence was now a repetition of (D, C, B, A), and a second one which
was simply a constant sequence, i.e.\ just 20 times the element A.

\begin{figure}[ht]
        \centering
        \includegraphics[width=0.47\textwidth]{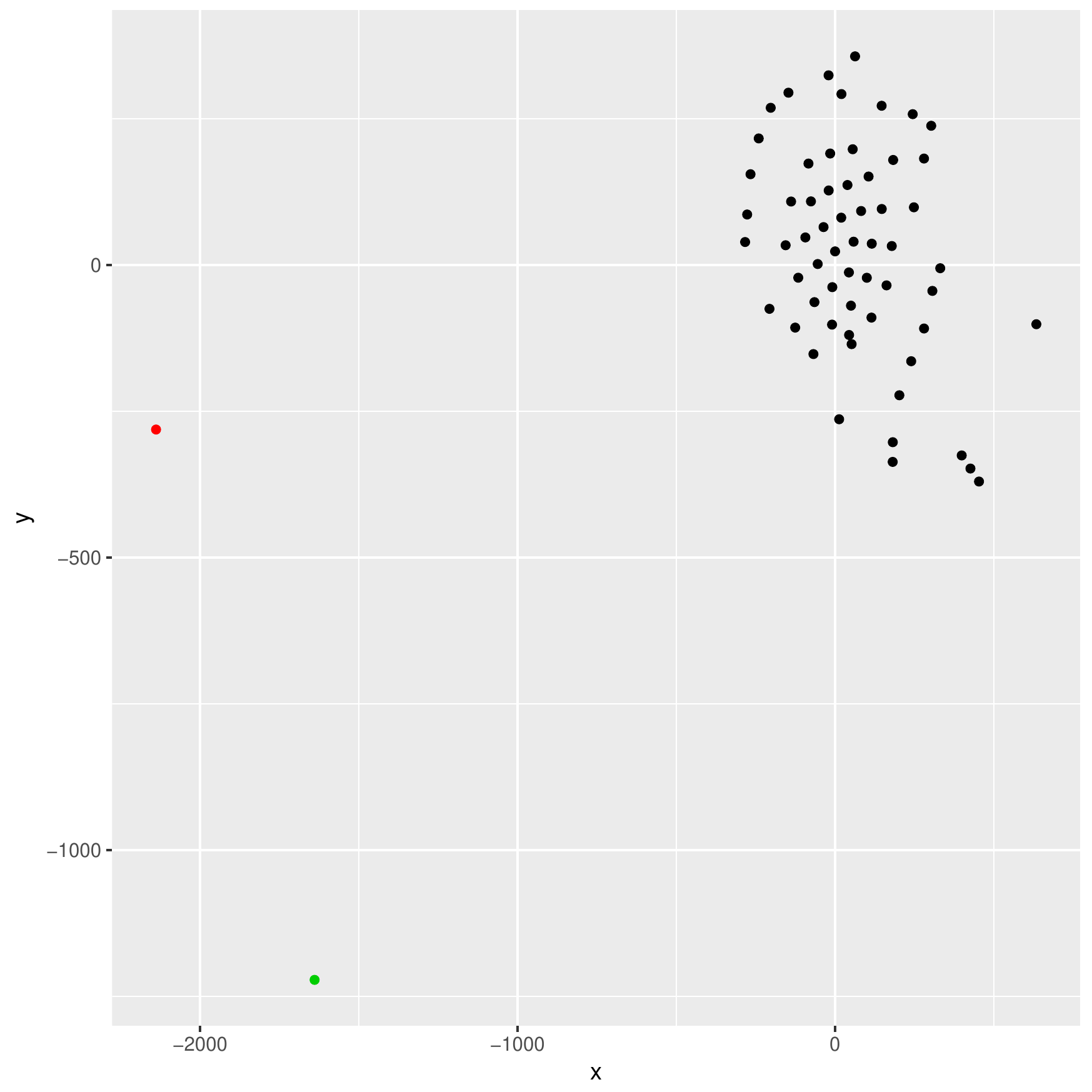}
        \caption{A 2D t-SNE plot of simulated data: 60 normal sequences in the
        upper right corner and two anomalous ones on the left and bottom left.}
        \label{fig:synthTwoAnoms}
\end{figure}

Then GLA was applied to this data, using 10 hidden states for the HMMs and ten
gauge sequences. The points in the feature space were then projected into
$\mathds R^2$ using t-SNE, see Figure \ref{fig:synthTwoAnoms}. The two outliers
are clearly visible near the left edge and the left lower edge of the plot.

Another experiment with simulated data is to test the capability of HMMs to
remember further back in time than just the previous event, i.e.\ that they are
more powerful than just Markov Chains. For this, we created
500 copies of the sequence $t_1$ and one single copy of sequence $t_2$:
\begin{equation}\begin{split}
 t_1 = &\mbox{(A, A, B, B, \ \ A, A, B, B, \ \ A, A, B, B,}\\
       &\mbox{A, A, B, B, \ \ A, A, B, B, \ \ A)}\\
 t_2 = &\mbox{(A, A, A, A, A, A, \ \ B, B, B, B, B, B,}\\
       &\mbox{A, B, A, B, A, B, A, B, A).}\\
\end{split}\end{equation}

\begin{figure}[ht]
        \centering
        \includegraphics[width=0.47\textwidth]{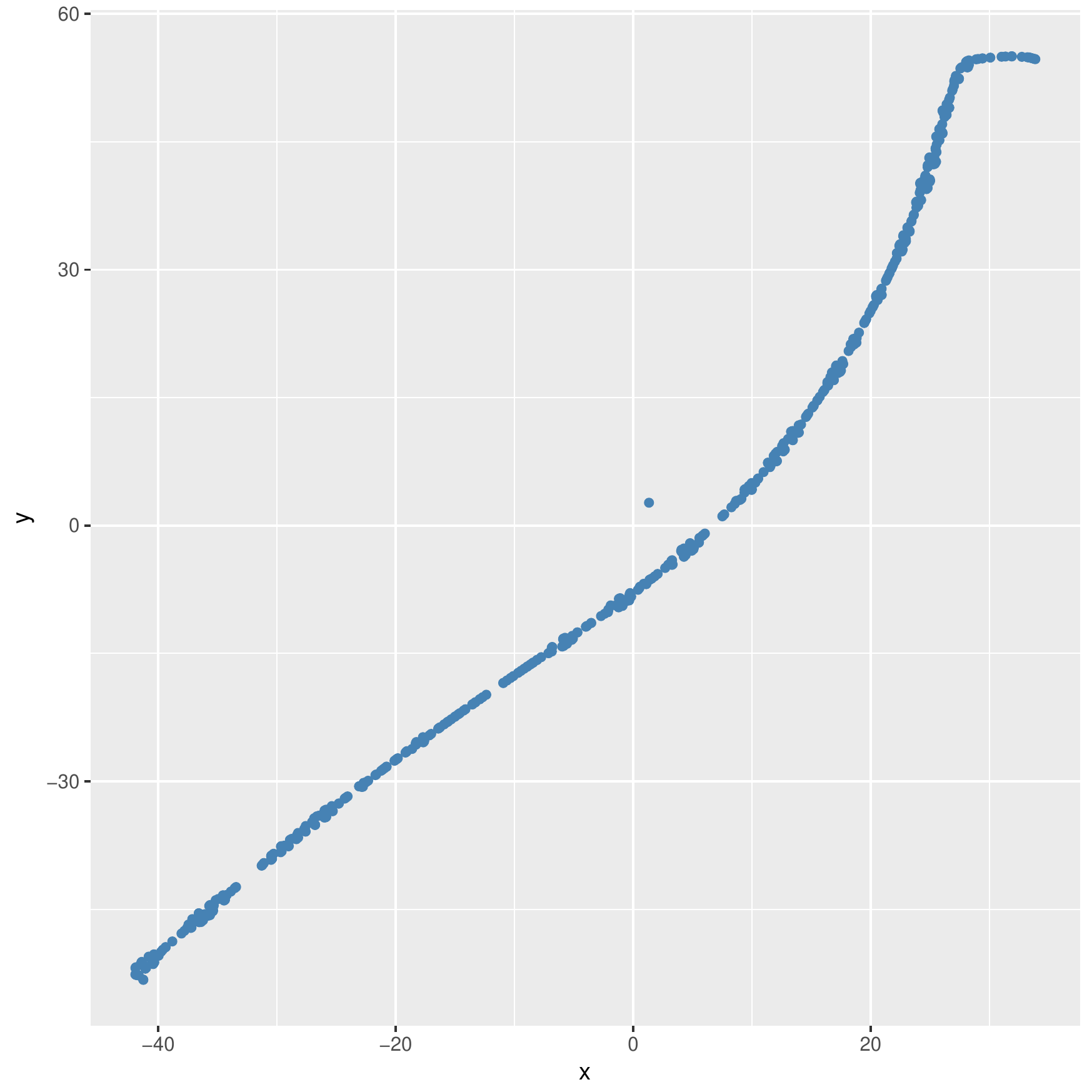}
        \caption{A 2D t-SNE plot of simulated data: 500 normal sequences in the
        long snake-shaped cluster and one anomalous one near the middle is
        clearly separated.}
        \label{fig:hmmMem}
\end{figure}

\noindent Learning a Markov chain on either of those two series would result in
the same transition matrix, namely the matrix with all entries equal to 0.5. So
a Markov chain model would not be capable to distinguish between those two time
series. However, both sequences have clearly different structure which should
be detectable for models with memory at least two steps back in the past. The
HMM models were configured with four states to be able to remember two steps
back. See Figure \ref{fig:hmmMem} for the result.

The image of the anomalous sequence is clearly visible as an outlier near the
middle of the figure.

\paragraph{Real data}
Since there is no standard benchmark data for anomaly detection, especially not
for syslog data, we used as real data syslog files collected from a Linux
laptop and examined those for anomalies. First, some preprocessing was done.
There are lots of syslog events that are of the same type and differ only in
some details like an IP address or a URL. Most of the time, those details are
not relevant and one would like to reduce all those variants into just one
single ``event type''. For example, a server might get connection requests from
thousands of clients and the belonging syslog events would only differ in the
URLs of the clients. We would like all those connection requests to be always
the same event type, just repeatedly logged.

One could understand this like a clustering of events where we are only
interested in the cluster id, which would then be the event type. This
clustering in itself is already a nontrivial task and there are several ways to
approach this problem, see for example \cite{vaarandi2003data} and
\cite{vaarandi2015logcluster}.

We used a simpler method: from each syslog line, the first three words of the
description field that are each at least three characters long and don't
contain any numbers or special characters are extracted as features. The
intuition behind this is that the irrelevant details usually show up more
towards the end of the log entry and there are rarely two events which coincide
in the first three non-numeric words and yet belong to different event types.
Verification on the data confirmed that this was a reasonable technique. Below,
this will be referred to as the ``three first words'' feature.

To evaluate GLA, we applied it to the syslog events of several applications.
Since our data was not labeled, we had to go through the sequences ourselves
and label the anomalies before evaluating our results.

The first example we will discuss here is for the application dhclient. The
data contains 46,757 dhclient events, each being one of 9 different event types
in the three first words feature. From this sequence, we extracted subsequences
over sliding windows of length 20 with shift distance 10. In our manual
inspection, we labeled 71 sequences as outliers which were due to rare
individual events they contained (we call those ``frequential anomalies''). But
we did not find any sequence that contained common events in a singular order
(we call those ``sequential anomalies'').

\begin{figure}[ht]
        \centering
        \includegraphics[width=0.47\textwidth]{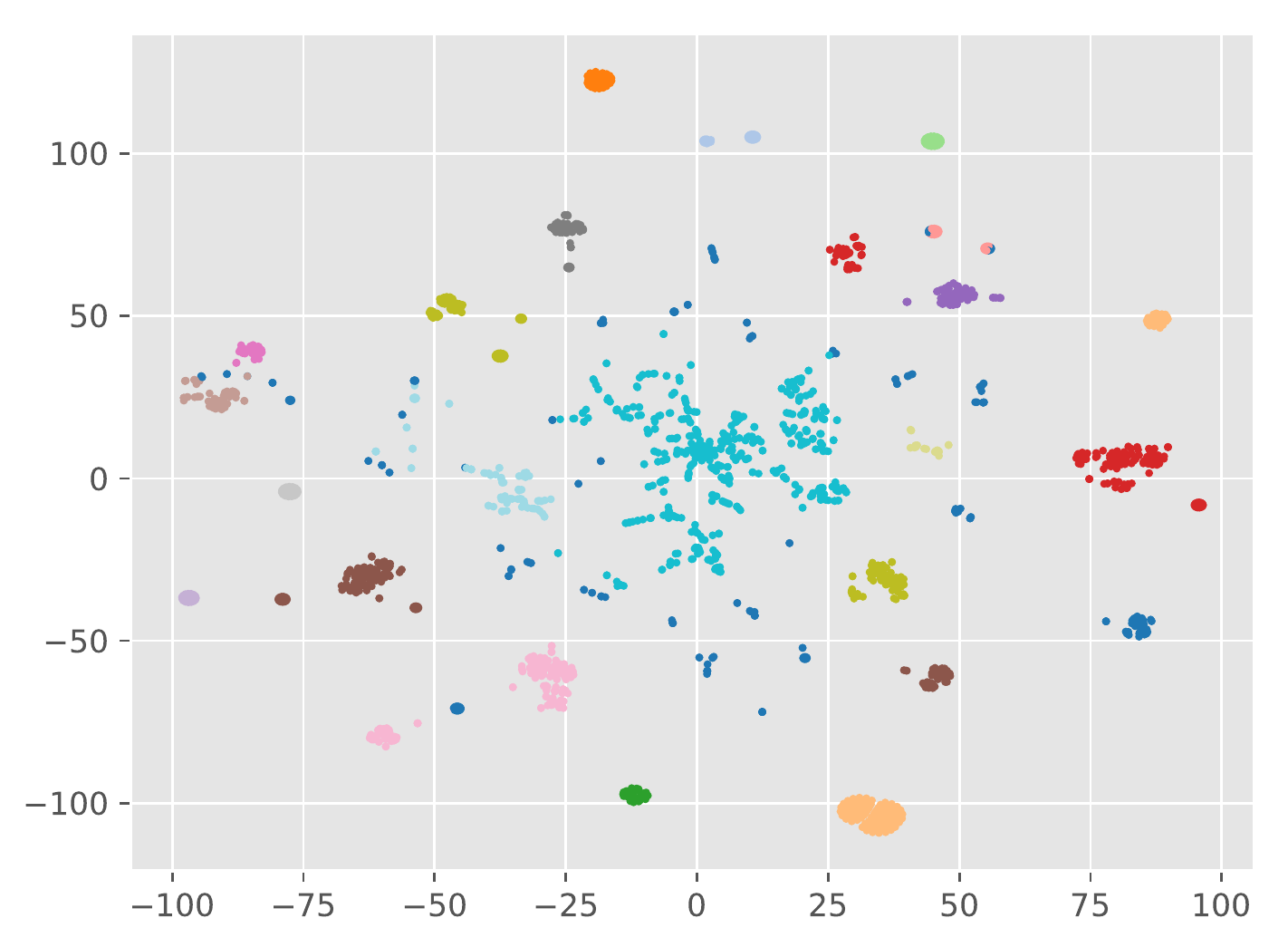}
        \caption{This shows the t-SNE plot of GLA for the application dhclient
        (sliding window size: 20, shift 10, possible symbols: 9, hidden states:
        20), clustered with HDBSCAN. The minimum cluster size parameter of
        HDBSCAN was set to 20. There are 51 outliers.}
        \label{fig:dhclient}
\end{figure}

On each of those dhclient subsequences, we trained an HMM with 20 hidden
states.  The clustered t-SNE plot of GLA is shown in Figure \ref{fig:dhclient}.
The clustering resulted in 51 outliers, all of them belonging to the 71 labeled
outliers.

\begin{figure}[ht]
        \centering
        \includegraphics[width=0.47\textwidth]{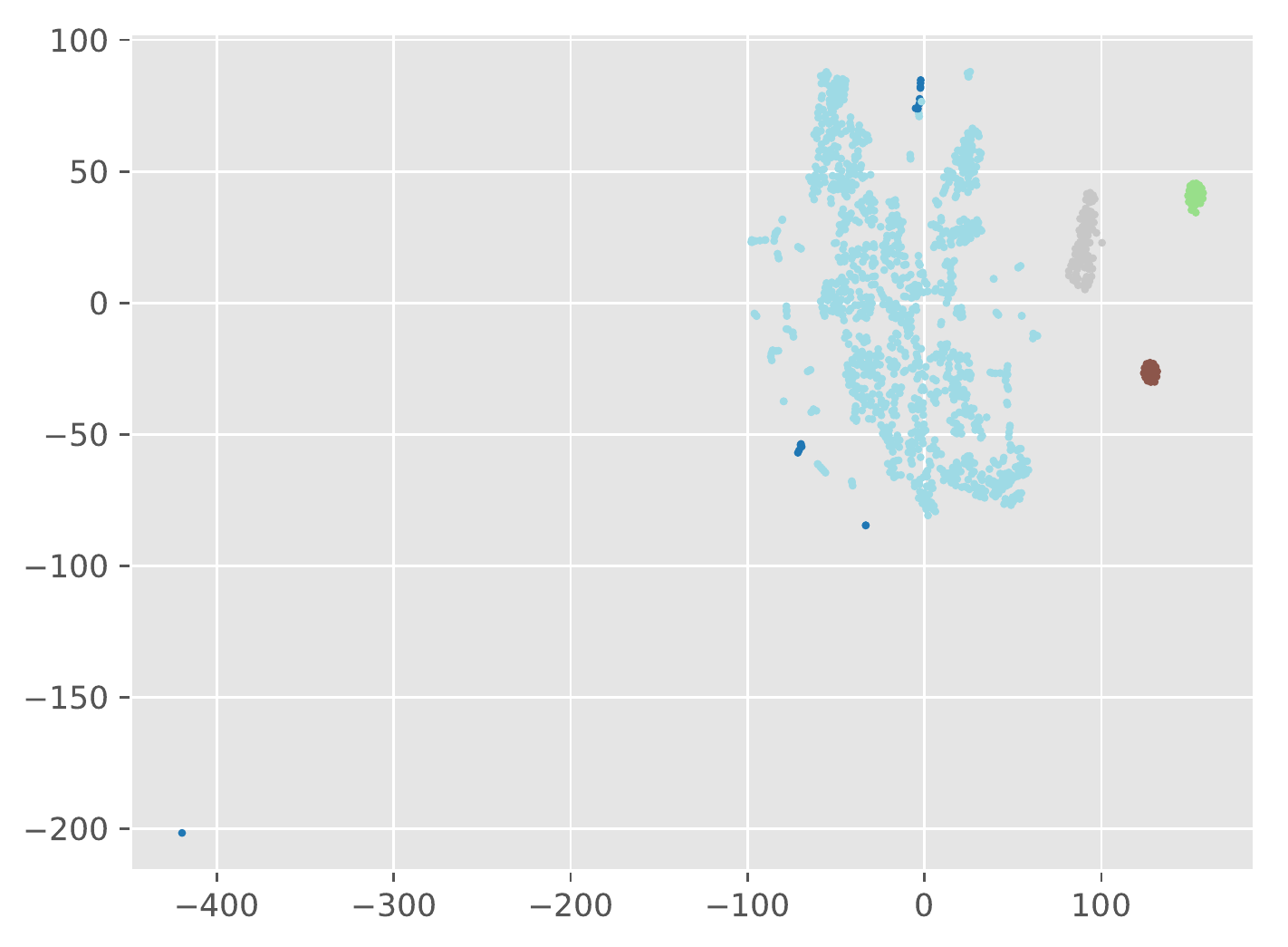}
        \caption{This shows the t-SNE plot of GLA for the application
        nm-dispatcher (sliding window size: 20, shift 10, possible symbols: 16,
        hidden states: 32), clustered with HDBSCAN. The minimum cluster size
        parameter of HDBSCAN was set to 20. There are 7 outliers.}
        \label{fig:nmDisp}
\end{figure}

The next example we discuss is for the application nm-dispatcher: the data
contained 25,352 such events, and again subsequences were extracted with a
sliding window of size 20 and a shift distance of 10.
The three first words feature for this application gives 16 different event
types. Manual labeling let to 34 outliers, consisting of 3 sequential and 31
frequential anomalies. Here, we trained HMMs with 32 hidden states. See Figure
\ref{fig:nmDisp} for the belonging plot.

GLA detected seven outliers, one from the sequential outliers, five from the
frequential outliers, and one unlabeled one. As usual, the very far outliers,
like the one here in the lower left corner, are frequential outliers.

\begin{figure}[ht]
        \centering
        \includegraphics[width=0.47\textwidth]{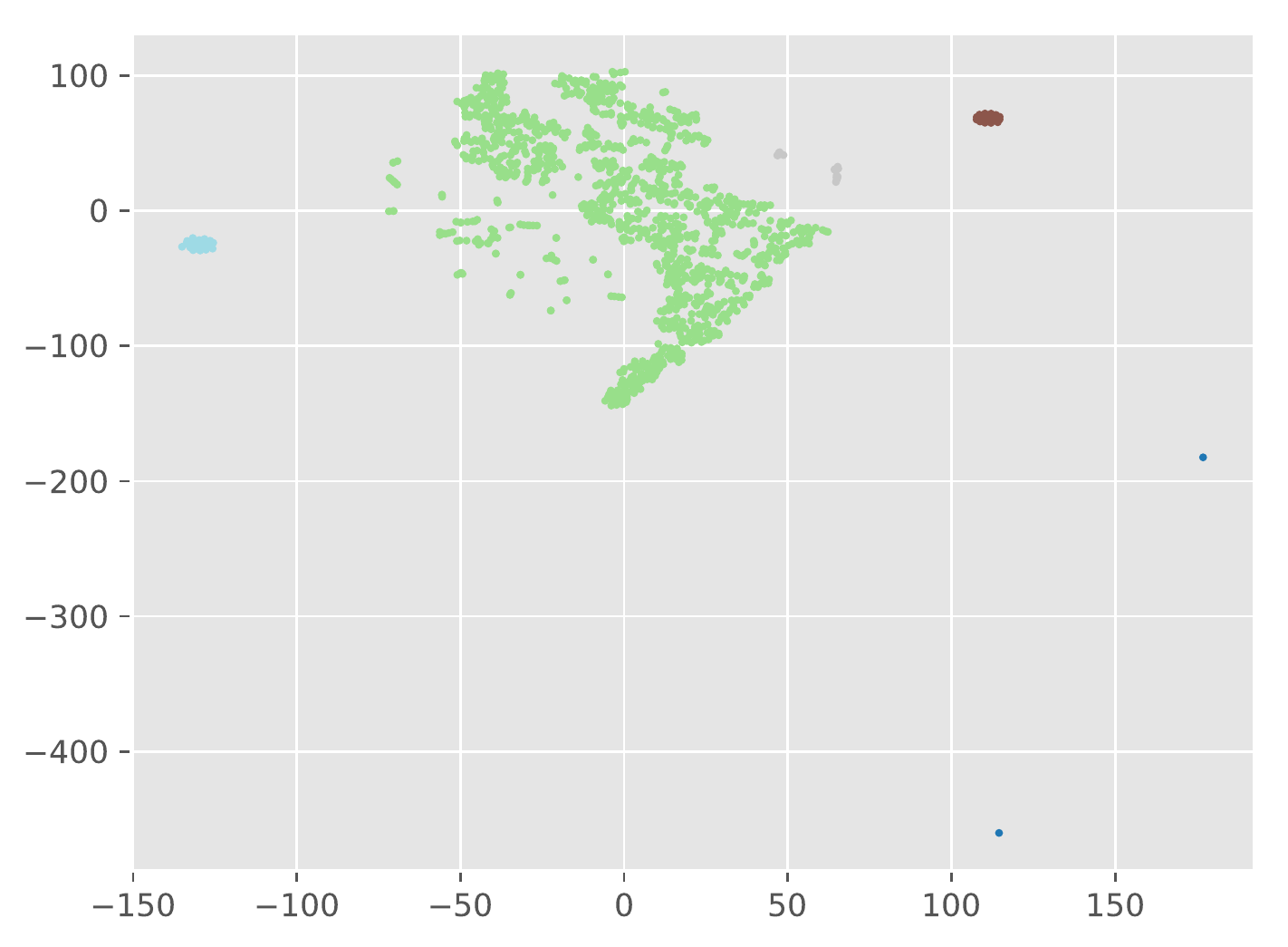}
        \caption{This shows the t-SNE plot of GLA for the application whoopsie
        (sliding window size: 20, shift 10, possible symbols: 15, hidden
        states: 32), clustered with HDBSCAN. The minimum cluster size parameter
        of HDBSCAN was set to 20. There are 2 outliers.}
        \label{fig:nmDisp2}
\end{figure}

Finally, we investigate the whoopsie application with one sequential outlier
and six frequential outliers. See Figure \ref{fig:nmDisp2} for the
visualization of GLA.

Two anomalies where detected. The outlier at the lower right is a frequential
outlier, while the one at the middle right is the sequential outlier.

The first five lines of the following table summarize the results.
\begin{center}
\begin{tabular}{cccc}
    \hline
               & dhclient     & nm-disp &  whoopsie \\
    \hline
    false pos. &       0      &             1 &         0 \\ [.2cm]
    false neg. &      20      &            28 &         5 \\ [.2cm]
    recall     &$\frac{51}{71} = 0.72$  &$\frac{6}{34} = 0.18$ &$\frac{2}{7} = 0.28$\\ [.2cm]
    precision  &$\frac{51}{51} = 1$  & $\frac{6}{7} = 0.86$ & $\frac{2}{2} = 1$  \\ [.2cm]
    \hline
    GLA: F1      &$0.84$ & $0.30$ & $0.44$  \\ [.2cm]
    \hline
    LSTM: F1     &$0.78$ & $0.31$ & $0.40$  \\ [.2cm]
\end{tabular}
\end{center}

Obviously, those numbers heavily depend on our choice of the minimum cluster
size of HDBSCAN. We preferred choices with less false positives. The parameter
was tuned for each application separately.

We compared those results with a popular deep learning anomaly detection
method, see \cite{malhotra2015long}. As discussed above, it uses LSTMs to
compute prediction probabilities and if those probabilities, evaluated for the
actual sequence, are smaller than a certain threshold, it is considered an
anomaly. We used an LSTM of length 20 with 256 units per cell, stacked four
times. This model was applied to the three event sequences described above.
For comparison with the results for GLA, the threshold for the prediction
probabilities was chosen to maximize the F1 score while producing the same
number of false positives as GLA. The F1 scores of LSTM have been entered into
the last row of the above table. We see that our GLA method outperforms the
LSTM model in the first and third case, while in the second case the results
are almost identical.

\section{Conclusions}
We have introduced GLA, a new method to detect anomalies in time series data.
It computes the HMMs for subsequences and then compares those HMMs with each
other by comparing their probability distributions. This is done by computing
the probabilities of those HMMs on a vector of gauge sequences and then
detecting outliers in the t-SNE projection using appropriate clustering
algorithms. The method has been successfully tested with Linux syslog data. The
experiments show that the method detects not only rare event anomalies but also
sequential anomalies.

\section{Future work}
In this paper, we have just described the first version of our method, the
proof of concept. There are several ways to improve on the current version.
The algorithm has still many parameters that need to be chosen, like the size
of the subsequences, the number of gauge sequences, the size of the gauge
sequences, the number of states of the HMMs, the outlier detection method, and
the belonging parameters of this method, such as the minimum cluster size in
the case of HDBSCAN. One should investigate to which extend the choice of those
parameters can be automated, maybe even learned.

Furthermore, it would be beneficial to see whether other projection methods
besides t-SNE, for example UMAP, could improve speed and outlier separation.
This should be investigated in tandem with a search for an optimal clustering
algorithm for this task.

A very interesting research topic would also be to compare different dynamic
Bayesian networks besides HMMs for capturing the sequential patterns in a given
population of time series.

\bibliographystyle{apalike}
{\small
\bibliography{adWithHmmGla}}

\end{document}